\pdfoutput=1

\documentclass[11pt]{article}

\usepackage{ACL2023}

\usepackage{times}
\usepackage{latexsym}

\usepackage{booktabs}
\usepackage{makecell}
\usepackage{multirow}
\usepackage{subcaption}
\usepackage{graphicx}
\usepackage{float}

\usepackage[T1]{fontenc}

\usepackage[utf8]{inputenc}

\usepackage{microtype}

\usepackage{inconsolata}

\title{Recipes for Sequential Pre-training of Multilingual \\ Encoder and Seq2Seq Models}

\author{Saleh Soltan\thanks{~~~Equal Contribution.} \\
  Alexa AI, Amazon\\
  \texttt{\normalsize ssoltan@amazon.com} \\\And
  Andy Rosenbaum\footnotemark[1] \\
  Alexa AI, Amazon \\
  \texttt{\normalsize andros@amazon.com} \\\And
  Tobias Falke \\
  Alexa AI, Amazon \\
  \texttt{\normalsize falket@amazon.de} \\\AND
  Qin Lu \\
  Alexa AI, Amazon \\
  \texttt{\normalsize luqn@amazon.com} \\\And
  Anna Rumshisky \\
  Alexa AI, Amazon \\
  Univ. of Massachusetts Lowell\\
  \texttt{\normalsize arrumshi@amazon.com} \\\And
  Wael Hamza \\
  Alexa AI, Amazon \\
  \texttt{\normalsize waelhamz@amazon.com} \\
 }

\begin{document}
\maketitle

\begin{abstract}

Pre-trained encoder-only and sequence-to-sequence (seq2seq)
models each have advantages,
however training both model types from scratch is computationally expensive.
We explore recipes to improve pre-training efficiency by initializing one model from the other.
(1) Extracting the encoder from a seq2seq model,
we show it
under-performs a Masked Language Modeling (MLM) encoder,
particularly on sequence labeling tasks.
Variations of masking during seq2seq training,
reducing the decoder size,
and continuing with a small amount of MLM
training do not close the gap.
(2) Conversely, using an encoder to warm-start seq2seq training,
we show that by unfreezing the encoder partway through training, we can match task performance of
a from-scratch seq2seq model. 
Overall, this two-stage approach is an efficient recipe
to obtain both a multilingual encoder and a seq2seq model,
matching the performance of training each model from scratch
while reducing the total compute cost by 27\%.

\end{abstract}

\section{Introduction and Related Work}

Transformer-based Pre-trained Language Models (PLMs) have become the main building blocks
when creating models for most Natural Language Processing (NLP) tasks.
PLMs come in three main architectures: decoder-only (e.g. GPT),
sequence-to-sequence (seq2seq, e.g. BART, T5), and encoder-only (e.g. BERT).
Multilingual models such as XLM-RoBERTa (encoder-only) and mBART/mT5 (seq2seq)
are also common.

\citet{Raffel2020ExploringTL} showed that seq2seq models can perform many NLP tasks on par with similarly-sized encoder-only models trained via Masked Language Modeling (MLM) by framing tasks such a sentence classification
or sequence labeling as text generation.
However, encoder models remain more efficient at inference for
sequence labeling tasks like Named Entity Recognition (NER)
and Part-of-Speech tagging (POS):
 an encoder can label all words in the sequence
 with a single forward pass, while a seq2seq model
 must generate each word's label autoregressively.

\begin{figure}[t]
    \centering
    \vspace{0.7cm}
    \includegraphics[width=0.48\textwidth]{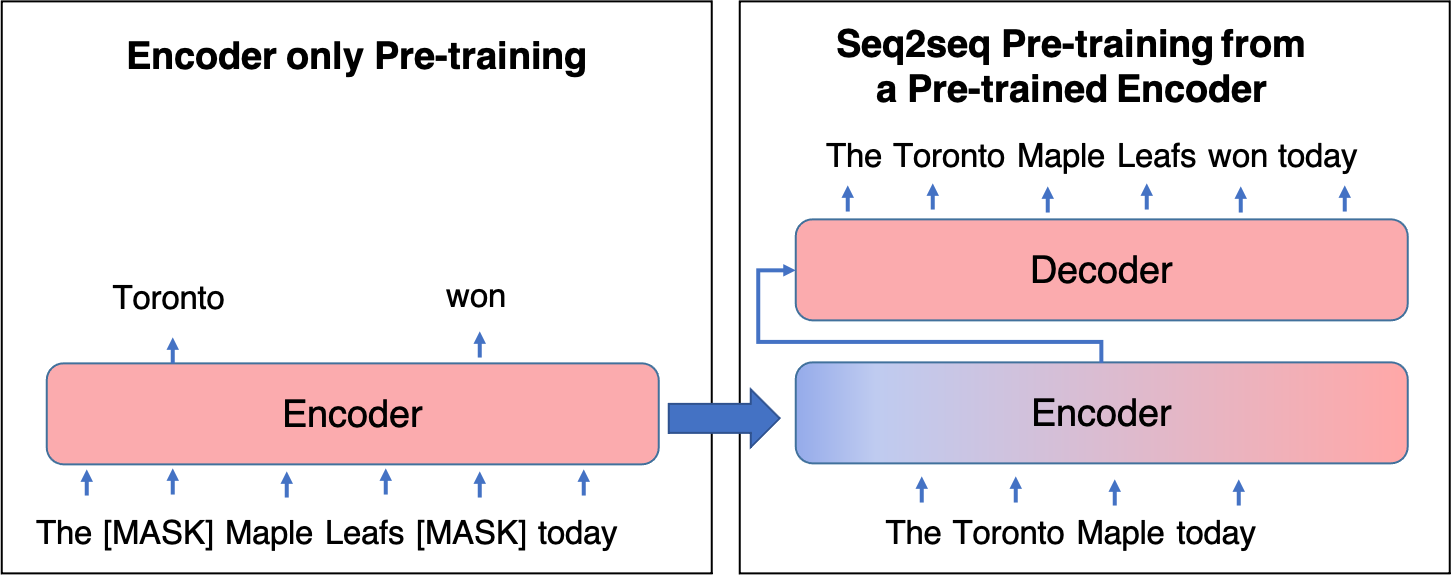}
    \caption{Two-stage seq2seq pre-training.
    First (left), we train the encoder via Masked Language Modeling (MLM).
    Second (right), we attach a randomly initialized decoder
    to the pre-trained MLM encoder, and train on the same data with
    de-noising objective.
    The encoder may remain frozen for part or all of the second stage.}
    \label{fig:two_stage_s2s}
\end{figure}

Motivated by the need for both an encoder model for efficient sequence labeling
and a seq2seq model for generative tasks like semantic parsing and summarization,
we explore recipes to pre-train both models.
Compared to training each model from scratch, we propose two
sequential training recipes which reduce the total compute cost
(Section \ref{sec:compute_cost_comparison}).

The first recipe is to extract the encoder of
a seq2seq model as proposed in \citet{ni-etal-2022-sentence}.
Although it performs well on classification tasks, we show that the encoder from seq2seq
under-performs a from-scratch encoder on sequence labeling tasks.
Variations of masking during seq2seq training
and reducing the decoder size do not provide a consistent benefit to the encoder.
We also explore continuing training the extracted encoder on MLM for a small number of updates.
However, we show it cannot consistently close the gap in performance across different datasets.

The second recipe is to warm-start seq2seq pre-training with an
encoder pre-trained via MLM (Figure~\ref{fig:two_stage_s2s}).
\citet{rothe-etal-2020-leveraging} proposed a similar idea for 
fine-tuning.
AlexaTM 20B and AlexaTM 5B
applied this recipe for pre-training, by warm-starting with Alexa Teacher Model encoders
\citep{Soltan2022AlexaTM2F,rosenbaum-etal-2022-linguist,FitzGerald2022AlexaTM}.
We add the novelty of comparing
to a seq2seq model pre-trained from scratch with the same data and codebase.
First, we observe that if the encoder is frozen the whole time,
the model under-performs a from-scratch seq2seq model on semantic parsing and summarization tasks.
While cross-attention fusion across different layers of the encoder reduces the performance gap,
we find that we can match performance of a from-scratch
model by using standard cross-attention and unfreezing the encoder
partway through training.

Overall, the second recipe demonstrates a viable approach for efficient pre-training of both a multilingual encoder and a multilingual seq2seq model,
matching the performance of training each model from
scratch, while using 27\% less total compute.

See Appendix \ref{sec:additional_related_work} for additional related work.

\section{Pre-Training Setup}

We describe our pre-training objectives,
models, datasets, two recipes for initializing one model type from the other,
and compare compute costs.

\subsection{Models}
\label{sec:models}

We pre-train ten models (Table \ref{table:models}):
one from-scratch encoder,
five from-scratch seq2seq models,
one encoder from a seq2seq model with continued MLM training,
and three two-stage seq2seq models warm-started with the from-scratch encoder.
We report the pre-training Compute Cost for each,
where ``TU'' (Training Units) is defined as
100k update steps for 12 model layers with hidden dimension 1024 and batch size 1M tokens
(Appendix \ref{sec:compute_cost_details}, \ref{sec:pre_train_details}).

\begin{table*}[th!]
\footnotesize
\begin{center}
\begin{tabular}{lcc|c|c|c}
\toprule
Model & 
\makecell{Encoder \\ Layers} & 
\makecell{Decoder \\ Layers} & 
\makecell{Encoder \\ Updates} & 
\makecell{Decoder \\ Updates}  & 
\makecell{Compute \\ Cost (TU)} \\ 
\midrule
\multicolumn{6}{c}{Encoder Model From Scratch (MLM only)} \\
\midrule
roberta-12e & 12 & 0 & 500k & 0 & 5.0\\
\midrule
\multicolumn{6}{c}{Seq2Seq Models From Scratch (de-noising only)} \\
\midrule
bart-12e12d      & 12 & 12 & 500k & 500k & 10.0 \\
bart-12e12d-mask & 12 & 12 & 500k & 500k & 10.0 \\
\midrule
bart-12e2d       & 12 & 2 & 500k & 500k & 5.8 \\
bart-12e2d-mask  & 12 & 2 & 500k & 500k & 5.8 \\
bart-12e1d-mask  & 12 & 1 & 500k & 500k & 5.4 \\
\midrule
\multicolumn{6}{c}{Recipe 1: Encoder of Seq2Seq + MLM} \\
\midrule
bart-12e12d+mlm        & 12 & 12 & 500k (s2s) + 100k & 500k & 10.0 (s2s) + 1.0 = 11.0 \\
\midrule
\multicolumn{6}{c}{Recipe 2: Two-Stage Seq2Seq Models (warm-start with MLM encoder)} \\
\midrule
2stage-bart-12e12d & 12 & 12 & 500k (MLM) & 500k & 5.0 (MLM) + 7.5 = 12.5 \\
2stage-bart-12e12d-attn-f & 12 & 12 & 500k (MLM) & 500k & 5.0 (MLM) + 7.5 = 12.5 \\
2stage-bart-12e12d-unfrz        & 12 & 12 & 500k (MLM) + 150k & 200k + 150k & 5.0 (MLM) + 6.0 = 11.0 \\
\bottomrule
\end{tabular}
\end{center}
\vspace{-0.2cm}
\caption{Model architecture details. All models use a batch size of 1M tokens with hidden dimension of 1024, feed-forward dimension of 4096 and 16 attention heads.
}
\vspace{-0.2cm}
\label{table:models}
\end{table*}

\subsubsection{Encoder Model From Scratch}
\label{sec:encoder_only_model}

We train an encoder model (``roberta-12e'' in Table~\ref{table:models}) following a similar recipe to
XLM-RoBERTa \citep{conneau-etal-2020-unsupervised},
using the MLM objectve (Figure~\ref{fig:mlm}) of
randomly masking 15\% of subword tokens,
as introduced in BERT \citep{devlin-etal-2019-bert}.
We use a batch size of 1M tokens and train for 500k update steps.
Notably, these settings match our seq2seq models.
We use ``PreLayerNorm'' \citep{Xiong2020OnLN}, moving the layer norms to
inside residual blocks to improve training stability.

\subsubsection{Seq2Seq Objectives}
Our seq2seq training follows the architecture and de-noising task of BART and mBART \citep{Lewis2020BARTDS,liu-etal-2020-multilingual-denoising}; the only architecture change we make is to
again use PreLayerNorm.

The de-noising objective selects
15\% of the tokens in the input (spans of length $\sim$ Poisson(3)),
and either (i) simply drops them, or (ii) replaces each selected span with a single mask token.
The model is trained to reconstruct the original input entirely.  See Figures \ref{fig:de_noising} and \ref{fig:de_noising_with_mask}, respectively. We add a suffix ``-mask'' to the model names that use masking instead of dropping the tokens. Intuitively, adding an explicit mask token for de-noising makes the reconstruction task easier,
as the decoder knows exactly where the missing tokens are needed.

\subsubsection{Seq2Seq Models From Scratch}

All of our seq2seq models use 12 \underline{e}ncoder layers (``12\underline{e}'').
The first five models are trained from scratch starting from randomly initialized weights.
The models ``bart-12e12\underline{d}'' and ``bart-12e12\underline{d}-mask'' use 12-layer \underline{d}ecoders
(same number as encoder layers) using the seq2seq de-noising training objective
without masking and with masking, respectively.
The remaining three models use a smaller decoder of either 2 layers
(``bart-12e2d'' without masking, ``bart-12e2d-mask'' with
masking) or 1 layer (``bart-12e1d-mask'', with masking).
We hypothesize that reducing the size of the decoder may strengthen the encoder
when it is extracted and used on its own.

\subsubsection{Recipe 1: Encoder of Seq2Seq + MLM}

We extract the encoder from the seq2seq model ``bart-12e12d''
and continue training via MLM for 100k updates (``bart-12e12d+mlm'').
We initialize the MLM head from the input embedding
and untie.

\subsubsection{Recipe 2: Two-Stage Seq2Seq Models}

Finally, we train three seq2seq models following the two-stage setup (Figure~\ref{fig:two_stage_s2s}).
We initialize the encoder weights of the seq2seq model with
the MLM encoder ``roberta-12e'' (Section \ref{sec:encoder_only_model})
and train via seq2seq de-noising without masking.
The first two models train for 500k updates
with the encoder always frozen:
``2stage-bart-12e12d'' uses standard cross-attention,
where the decoder attends to only the final encoder layer,
and ``2stage-bart-12e12d-\underline{attn-f}''
uses a novel application of \textbf{attention fusion} \citep{cao-etal-2022-attention}
during cross-attention, where the decoder
attends to all encoder layers.

The last model, ``2stage-bart-12e12d-\underline{unfrz}''
uses standard cross-attention and
\textbf{unfreezes the encoder partway through training}, applying
200k update steps with the encoder frozen, then 150k update steps with the encoder unfrozen.

In all cases, we initialize and tie the decoder embeddings from/to the encoder embeddings and
keep them frozen as long as the encoder is frozen.
The LM head is also initialized from the encoder embeddings, but it is untied from the embeddings and unfrozen from the beginning of the training.

\subsubsection{Compute Cost Comparison}
\label{sec:compute_cost_comparison}

The baseline of training both models from scratch has a compute cost of 15.0 TU:
5.0 TU for ``roberta-12e'' plus 10.0 TU for ``bart-12e12d''.
Our proposed recipes reduce the total compute cost
either by 17\% (to 12.5 TU) or by 27\% (to 11.0 TU).

\subsection{Pretraining Dataset}
We pre-train on a combination of Wikipedia and mC4 \citep{xue-etal-2021-mt5} data in 12 languages: Arabic, English, French, German, Hindi, Italian, Japanese, Marathi, Portuguese, Spanish, Tamil, and Telugu. We pack sequences of tokens to produce sequences of approximately 512 subword units. We allow unrelated content to be packed together in the same sequence, separated with a special symbol ``[DOC]''. Maintaining a relatively constant number of subword sequences reduces padding and results in efficient compute. We up-sample data for different languages following~\citet{conneau-etal-2020-unsupervised}.

\section{Fine-Tuning Results}
\label{sec:ft_res}

\begin{table*}
\footnotesize
\setlength{\tabcolsep}{2.7pt}
\begin{center}
\begin{tabular}{l|cc|cc|cc|cc|cc}
\toprule
& \multicolumn{4}{c|}{Classification} & \multicolumn{6}{c}{Sequence Labeling} \\
\midrule
\multirow{3}{*}{Encoder} & \multicolumn{2}{c|}{XNLI (acc.)} & \multicolumn{2}{c|}{mATIS++ IC (acc.)} & \multicolumn{2}{c|}{mATIS++ SL (f1)} & \multicolumn{2}{c|}{WikiANN (f1)} & \multicolumn{2}{c}{UDPOS (f1)} \\
& en & avg-0s & en & avg-0s & en & avg-0s & en & avg-0s & en & avg-0s\\ 
\midrule
\multicolumn{11}{c}{Encoder Model From Scratch (MLM only)} \\
\midrule
roberta-12e       & $\textbf{84.5}_{\pm{}0.5}$ & $\textbf{75.8}_{\pm{}0.2}$ & $\textbf{97.8}_{\pm{}0.1}$ & $87.2_{\pm{}4.1}$ & $\textbf{95.7}_{\pm{}0.1}$ & $\textbf{61.6}_{\pm{}0.6}$ & $\textbf{83.0}_{\pm{}0.1}$ & $\textbf{61.1}_{\pm{}0.4}$ & $\textbf{95.8}_{\pm{}0.0}$ & $\textbf{73.5}_{\pm{}0.2}$\\
\midrule
 \multicolumn{11}{c}{Encoder of Seq2Seq Models (de-noising only)} \\
 \midrule
 bart-12e12d       & $\underline{83.9}_{\pm{}0.2}$ & $74.7_{\pm{}0.3}$ & $96.8_{\pm{}0.1}$ & $86.2_{\pm{}1.5}$ & $92.5_{\pm{}0.3}$ & $44.3_{\pm{}1.3}$ & $76.6_{\pm{}0.2}$ & $52.1_{\pm{}0.9}$ & $94.3_{\pm{}0.7}$ & $\underline{61.5}_{\pm{}0.4}$ \\
 bart-12e12d-mask & $\underline{83.9}_{\pm{}0.4}$ & $\underline{75.0}_{\pm{}0.6}$ & $97.1_{\pm{}0.1}$ & $87.3_{\pm{}0.7}$ & $91.1_{\pm{}0.9}$ & $41.3_{\pm{}1.3}$ & $73.2_{\pm{}0.1}$ & $48.4_{\pm{}0.6}$ & $93.3_{\pm{}0.1}$ & $55.1_{\pm{}0.4}$ \\
 \midrule
 bart-12e2d        & $71.3_{\pm{}0.1}$ & $59.7_{\pm{}0.5}$ & $96.1_{\pm{}0.1}$ & $79.1_{\pm{}0.8}$ & $91.4_{\pm{}0.1}$ & $38.2_{\pm{}1.7}$ & $69.3_{\pm{}0.5}$ & $42.9_{\pm{}0.1}$ & $92.1_{\pm{}0.1}$ & $50.7_{\pm{}0.5}$ \\
 bart-12e2d-mask   & $82.9_{\pm{}0.3}$ & $73.8_{\pm{}0.2}$ & $96.8_{\pm{}0.1}$ & $\textbf{88.1}_{\pm{}0.9}$ & $92.3_{\pm{}0.3}$ & $48.0_{\pm{}1.4}$ & $76.5_{\pm{}0.2}$ & $\underline{54.0}_{\pm{}0.6}$ & $93.3_{\pm{}0.1}$ & $54.0_{\pm{}0.6}$ \\
 bart-12e1d-mask   & $82.4_{\pm{}0.2}$ & $72.7_{\pm{}0.1}$ & $97.0_{\pm{}0.1}$ & $\underline{87.6}_{\pm{}0.5}$ & $92.8_{\pm{}0.5}$ & $49.3_{\pm{}1.2}$ & $74.6_{\pm{}0.5}$ & $48.5_{\pm{}0.3}$ & $92.4_{\pm{}0.1}$ & $46.3_{\pm{}1.7}$ \\
\midrule
\multicolumn{11}{c}{Recipe 1: Encoder of Seq2Seq Model + MLM} \\
\midrule
 bart-12e12d+mlm  & $80.3_{\pm{}0.4}$ & $69.0_{\pm{}0.4}$ & $\underline{97.2}_{\pm{}0.4}$ & $83.9_{\pm{}1.6}$ & $\underline{95.3}_{\pm{}0.2}$ & $\underline{56.5}_{\pm{}2.8}$ & $\underline{79.9}_{\pm{}0.2}$ & $47.5_{\pm{}0.5}$ & $\underline{95.1}_{\pm{}0.0}$ & $60.7_{\pm{}0.9}$ \\
\bottomrule
\end{tabular}
\end{center}
\vspace{-0.2cm}
\caption{Encoder results per task, English and avg. zero-shot. The best (second) mean result is bolded (underlined).
}
\label{table:enc_all}
\end{table*}

\begin{table*}[th!]
\footnotesize
\begin{center}
\begin{tabular}{l | cc | ccc}
\toprule
\multirow{2}{*}{Seq2Seq Models} & \multicolumn{2}{c|}{mTOP (acc.)} & \multicolumn{3}{c}{XSUM (ROUGE)} \\
& en & avg-0s & R-1 & R-2 & R-L \\ 
\midrule
\multicolumn{6}{c}{Seq2Seq Models From Scratch (de-noising only)} \\
\midrule
 bart-12e12d                & $\textbf{83.4}_{\pm0.2}$ & $45.7_{\pm1.1}$ & $\underline{40.37}_{\pm0.07}$ & $17.37_{\pm0.06}$ & $32.46_{\pm0.06}$ \\
 bart-12e12d-mask           & $83.2_{\pm0.5}$ & $\underline{46.9}_{\pm0.5}$ & $\textbf{40.63}_ {\pm0.09}$ & $\underline{17.48}_{\pm0.10}$ & $\underline{32.63}_{\pm0.06}$ \\
 \midrule
 \multicolumn{6}{c}{Recipe 2: Two-Stage Seq2Seq Models (warm-start with MLM encoder)} \\
\midrule
 2stage-bart-12e12d         & $82.0_{\pm1.1}$ & $46.8_{\pm1.1}$ & $40.12_{\pm0.06}$ & $17.13_{\pm0.03}$ & $32.16_{\pm0.01}$ \\
  2stage-bart-12e12d-attn-f  & $80.6_{\pm1.3}$ & $46.4_{\pm0.5}$ & $40.13_{\pm0.06}$ & $17.24_{\pm0.07}$ & $32.28_{\pm0.03}$ \\
 2stage-bart-12e12d-unfrz         & $\underline{83.3}_{\pm0.2}$ & $\textbf{48.2}_{\pm0.5}$ & $\textbf{40.63}_{\pm0.11}$ & $\textbf{17.58}_{\pm0.03}$ & $\textbf{32.65}_{\pm0.05}$  \\
\bottomrule
\end{tabular}
\end{center}
\vspace{-0.2cm}
\caption{Seq2Seq results on mTOP cross-lingual semantic parsing and XSUM English summarization.}
\vspace{-0.2cm}
\label{table:generative}
\end{table*}

We present the results on fine-tuning our pre-trained models.
All runs are averaged over three random seeds and reported as mean $\pm$ standard deviation.
See Appendix \ref{sec:ft_hyperparams} for hyperparameters.

\subsection{Encoder Model Results}

In Table \ref{table:enc_all}, we compare the performance of
our encoder models
on four datasets:
(1) XNLI \citep{conneau-etal-2018-xnli} sentence-pair classification,
(2) mATIS++ \citep{xu-etal-2020-end} joint Intent Classification (IC) and Slot Labeling (SL),
(3) WikiANN \citep{pan-etal-2017-cross} token-level Named Entity Recognition (NER), and
(4) UDPOS \citep{Nivre2020UniversalDV}
token-level Part-of-Speech tagging (POS) (XTREME \citep{pmlr-v119-hu20b} version).
For each task, we follow the cross-lingual zero-shot setting:
train and validate on English data only, then report on the test set
in English (``en'') and the average over the zero-shot langauges (``avg-0s'').
Appendix \ref{sec:res_all_langs} shows results on each language.

\textbf{We find that the MLM encoder performs best on all tasks}
except for mATIS++ IC avg-0s setting.
The encoder of seq2seq (``bart-12e12d'')
is only slightly behind on the sentence-level tasks, on en/avg-0s by
0.6/1.1 points on XNLI (83.9 vs. 84.5 / 74.7 vs. 75.8), and
1.0/1.0 points on mATIS++ IC (96.8 vs. 97.8 / 86.2 vs. 87.2).
However, \textbf{the gap is much larger on the sequence labeling tasks}:
on en/avg-0s,
3.2/17.3 points on mATIS++ SL (92.5 vs. 95.7 / 44.3 vs. 61.6),
6.4/9.0 points on WikiANN NER (76.6 vs. 83.0 / 52.1 vs. 61.1), and
1.5/12.0 on UDPOS (94.3 vs. 95.8 / 61.5 vs. 73.5).
This suggests that seq2seq pre-training may give the encoder
the knowledge to perform sentence-level tasks, while MLM pre-training
may be particularly effective for sequence labeling tasks which use the token-level representations directly.

With a 12-layer decoder, the explicit mask token during 
seq2seq pre-training does not seem to improve the encoder.
However, when the decoder has only 2 layers, the mask
token is crucial: ``bart-12e2d-mask'' out-performs ``bart-12e2d'' by a wide margin across tasks.
We hypothesize that the mask token
makes de-noising easier, by signaling
where tokens should be filled in, and without this signal,
the task is too challenging for a seq2seq model with just a 2-layer decoder.
Reducing the decoder further to only 1 layer
does not benefit the encoder.

Continuing training the seq2seq-extracted encoder on MLM for 100k updates
does not close the gap to the from-scratch encoder across datasets.
Some tasks improve, while others degrade.

\subsection{Seq2Seq Model Results}

We evaluate the generation quality of our seq2seq models
on two datasets: mTOP \citep{li-etal-2021-mtop} cross-lingual zero-shot semantic parsing,
and XSUM \citep{narayan-etal-2018-dont} English summarization.
For mTOP, following CLASP \citep{rosenbaum-etal-2022-clasp}, we use space-joined tokens as input, word sentinels,
and SCIEM (Space- and Case-Insensitive Exact Match) metric.
For both datasets, we generate outputs using beam search with k=3.

As shown in Table~\ref{table:generative},
\textbf{the two-stage model with encoder unfrozen partway through training
is on-par with the from-scratch seq2seq model}:
compared to ``bart-12e12d '', ``2stage-bart-12e12d-unfrz''
is only 0.1 points behind on mTOP en (83.3 vs. 83.4) yet 2.5
points ahead on cross-lingual zero-shot (48.2 vs. 45.7).
On XSUM, the two-stage model is on-par or slightly better
than the from-scratch seq2seq models.

Masking during seq2seq pre-training does not greatly impact generation quality.
When the encoder is frozen (``2stage-bart-12e12d''), the results are slightly behind;
attention fusion (``2stage-bart-12e12d-attn-f'') does not provide a clear benefit.

Overall,
our proposed
\textbf{two-stage seq2seq pre-training recipe provides both a multilingual
encoder and a seq2seq model
on-par with the two models trained from scratch,
while reducing compute cost by 27\%} (from 15.0 to 11.0 TU).

\section{Conclusion and Future Work}

In this work, we studied recipes to efficiently pre-train both a multilingual encoder and a seq2seq model by re-using the weights from one model for the other. We found that the most effective recipe is to start training of a seq2seq model from a pre-trained encoder and unfreeze it partway through the training.
Future work can explore even more efficient pre-training strategies such as jointly training on MLM and sequence-level de-noising objectives, and probe further why the encoders trained as part of a seq2seq model do not do well on sequence labeling tasks.

\section{Limitations}

Our proposed two-stage training recipe is beneficial under the assumption that a pre-trained model is needed for generative as well as sequence labeling tasks. We believe that is typically the case, as one tries to offset the pre-training investment by using the model for as many tasks as possible, but this assumption might not apply in all cases. 
While we assess the effect of randomness on fine-tuning results by using multiple seeds, we have not done that for the pre-training itself. Even at our medium-size scale, it is already prohibitively expensive to do so. 
The evidence for the effectiveness of the two-stage approach is also limited by the number of tasks evaluated (2 sequence classification tasks, 2 sequence labeling tasks, 2 generation tasks), but we believe it is a reasonable trade-off between robust results and compute investment.

\section{Acknowledgments}
We thank Kai-Wei Chang, Nicolas Guenon Des Mesnards, and the anonymous ACL reviewers for their helpful feedback on our work.

\bibliography{anthology,custom}

\begin{thebibliography}{40}
\expandafter\ifx\csname natexlab\endcsname\relax\def\natexlab#1{#1}\fi

\bibitem[{Bao et~al.(2020)Bao, Dong, Wei, Wang, Yang, Liu, Wang, Piao, Gao,
  Zhou, and Hon}]{Bao2020UniLMv2PL}
Hangbo Bao, Li~Dong, Furu Wei, Wenhui Wang, Nan Yang, Xiaodong Liu, Yu~Wang,
  Songhao Piao, Jianfeng Gao, Ming Zhou, and Hsiao-Wuen Hon. 2020.
\newblock Unilmv2: Pseudo-masked language models for unified language model
  pre-training.
\newblock \emph{ArXiv}, abs/2002.12804.

\bibitem[{Brown et~al.(2020)Brown, Mann, Ryder, Subbiah, Kaplan, Dhariwal,
  Neelakantan, Shyam, Sastry, Askell, Agarwal, Herbert-Voss, Krueger, Henighan,
  Child, Ramesh, Ziegler, Wu, Winter, Hesse, Chen, Sigler, Litwin, Gray, Chess,
  Clark, Berner, McCandlish, Radford, Sutskever, and
  Amodei}]{Brown2020LanguageMA}
Tom~B. Brown, Benjamin Mann, Nick Ryder, Melanie Subbiah, Jared Kaplan,
  Prafulla Dhariwal, Arvind Neelakantan, Pranav Shyam, Girish Sastry, Amanda
  Askell, Sandhini Agarwal, Ariel Herbert-Voss, Gretchen Krueger, T.~J.
  Henighan, Rewon Child, Aditya Ramesh, Daniel~M. Ziegler, Jeff Wu, Clemens
  Winter, Christopher Hesse, Mark Chen, Eric Sigler, Mateusz Litwin, Scott
  Gray, Benjamin Chess, Jack Clark, Christopher Berner, Sam McCandlish, Alec
  Radford, Ilya Sutskever, and Dario Amodei. 2020.
\newblock Language models are few-shot learners.
\newblock \emph{ArXiv}, abs/2005.14165.

\bibitem[{Cao et~al.(2022)Cao, Satya~Prakash, and
  Hamza}]{cao-etal-2022-attention}
Jin Cao, Chandana Satya~Prakash, and Wael Hamza. 2022.
\newblock \href {https://doi.org/10.18653/v1/2022.findings-naacl.64} {Attention
  fusion: a light yet efficient late fusion mechanism for task adaptation in
  {NLU}}.
\newblock In \emph{Findings of the Association for Computational Linguistics:
  NAACL 2022}, pages 857--866, Seattle, United States. Association for
  Computational Linguistics.

\bibitem[{Chen et~al.(2022)Chen, Yin, Shang, Jiang, Qin, Wang, Wang, Chen, Liu,
  and Liu}]{chen-etal-2022-bert2bert}
Cheng Chen, Yichun Yin, Lifeng Shang, Xin Jiang, Yujia Qin, Fengyu Wang, Zhi
  Wang, Xiao Chen, Zhiyuan Liu, and Qun Liu. 2022.
\newblock \href {https://doi.org/10.18653/v1/2022.acl-long.151} {bert2{BERT}:
  Towards reusable pretrained language models}.
\newblock In \emph{Proceedings of the 60th Annual Meeting of the Association
  for Computational Linguistics (Volume 1: Long Papers)}, pages 2134--2148,
  Dublin, Ireland. Association for Computational Linguistics.

\bibitem[{Chen et~al.(2019)Chen, Zhuo, and Wang}]{chen2019bert}
Qian Chen, Zhu Zhuo, and Wen Wang. 2019.
\newblock \href {http://arxiv.org/abs/1902.10909} {Bert for joint intent
  classification and slot filling}.

\bibitem[{Chowdhery et~al.(2022)Chowdhery, Narang, Devlin, Bosma, Mishra,
  Roberts, Barham, Chung, Sutton, Gehrmann, Schuh, Shi, Tsvyashchenko, Maynez,
  Rao, Barnes, Tay, Shazeer, Prabhakaran, Reif, Du, Hutchinson, Pope, Bradbury,
  Austin, Isard, Gur-Ari, Yin, Duke, Levskaya, Ghemawat, Dev, Michalewski,
  Garc{\'i}a, Misra, Robinson, Fedus, Zhou, Ippolito, Luan, Lim, Zoph,
  Spiridonov, Sepassi, Dohan, Agrawal, Omernick, Dai, Pillai, Pellat,
  Lewkowycz, Moreira, Child, Polozov, Lee, Zhou, Wang, Saeta, Diaz, Firat,
  Catasta, Wei, Meier-Hellstern, Eck, Dean, Petrov, and
  Fiedel}]{Chowdhery2022PaLMSL}
Aakanksha Chowdhery, Sharan Narang, Jacob Devlin, Maarten Bosma, Gaurav Mishra,
  Adam Roberts, Paul Barham, Hyung~Won Chung, Charles Sutton, Sebastian
  Gehrmann, Parker Schuh, Kensen Shi, Sasha Tsvyashchenko, Joshua Maynez,
  Abhishek~Baindoor Rao, Parker Barnes, Yi~Tay, Noam~M. Shazeer, Vinodkumar
  Prabhakaran, Emily Reif, Nan Du, Benton~C. Hutchinson, Reiner Pope, James
  Bradbury, Jacob Austin, Michael Isard, Guy Gur-Ari, Pengcheng Yin, Toju Duke,
  Anselm Levskaya, Sanjay Ghemawat, Sunipa Dev, Henryk Michalewski, Xavier
  Garc{\'i}a, Vedant Misra, Kevin Robinson, Liam Fedus, Denny Zhou, Daphne
  Ippolito, David Luan, Hyeontaek Lim, Barret Zoph, Alexander Spiridonov, Ryan
  Sepassi, David Dohan, Shivani Agrawal, Mark Omernick, Andrew~M. Dai,
  Thanumalayan~Sankaranarayana Pillai, Marie Pellat, Aitor Lewkowycz,
  Erica~Oliveira Moreira, Rewon Child, Oleksandr Polozov, Katherine Lee,
  Zongwei Zhou, Xuezhi Wang, Brennan Saeta, Mark Diaz, Orhan Firat, Michele
  Catasta, Jason Wei, Kathleen~S. Meier-Hellstern, Douglas Eck, Jeff Dean, Slav
  Petrov, and Noah Fiedel. 2022.
\newblock {PaLM}: Scaling language modeling with pathways.
\newblock \emph{ArXiv}, abs/2204.02311.

\bibitem[{Conneau et~al.(2020)Conneau, Khandelwal, Goyal, Chaudhary, Wenzek,
  Guzm{\'a}n, Grave, Ott, Zettlemoyer, and
  Stoyanov}]{conneau-etal-2020-unsupervised}
Alexis Conneau, Kartikay Khandelwal, Naman Goyal, Vishrav Chaudhary, Guillaume
  Wenzek, Francisco Guzm{\'a}n, Edouard Grave, Myle Ott, Luke Zettlemoyer, and
  Veselin Stoyanov. 2020.
\newblock \href {https://doi.org/10.18653/v1/2020.acl-main.747} {Unsupervised
  cross-lingual representation learning at scale}.
\newblock In \emph{Proceedings of the 58th Annual Meeting of the Association
  for Computational Linguistics}, pages 8440--8451, Online. Association for
  Computational Linguistics.

\bibitem[{Conneau et~al.(2018)Conneau, Rinott, Lample, Williams, Bowman,
  Schwenk, and Stoyanov}]{conneau-etal-2018-xnli}
Alexis Conneau, Ruty Rinott, Guillaume Lample, Adina Williams, Samuel Bowman,
  Holger Schwenk, and Veselin Stoyanov. 2018.
\newblock \href {https://doi.org/10.18653/v1/D18-1269} {{XNLI}: Evaluating
  cross-lingual sentence representations}.
\newblock In \emph{Proceedings of the 2018 Conference on Empirical Methods in
  Natural Language Processing}, pages 2475--2485, Brussels, Belgium.
  Association for Computational Linguistics.

\bibitem[{Devlin et~al.(2019)Devlin, Chang, Lee, and
  Toutanova}]{devlin-etal-2019-bert}
Jacob Devlin, Ming-Wei Chang, Kenton Lee, and Kristina Toutanova. 2019.
\newblock \href {https://doi.org/10.18653/v1/N19-1423} {{BERT}: Pre-training of
  deep bidirectional transformers for language understanding}.
\newblock In \emph{Proceedings of the 2019 Conference of the North {A}merican
  Chapter of the Association for Computational Linguistics: Human Language
  Technologies, Volume 1 (Long and Short Papers)}, pages 4171--4186,
  Minneapolis, Minnesota. Association for Computational Linguistics.

\bibitem[{Dong et~al.(2019)Dong, Yang, Wang, Wei, Liu, Wang, Gao, Zhou, and
  Hon}]{Dong2019UnifiedLM}
Li~Dong, Nan Yang, Wenhui Wang, Furu Wei, Xiaodong Liu, Yu~Wang, Jianfeng Gao,
  M.~Zhou, and Hsiao-Wuen Hon. 2019.
\newblock Unified language model pre-training for natural language
  understanding and generation.
\newblock \emph{ArXiv}, abs/1905.03197.

\bibitem[{FitzGerald et~al.(2022)FitzGerald, Ananthakrishnan, Arkoudas,
  Bernardi, Bhagia, Delli~Bovi, Cao, Chada, Chauhan, Chen, Dwarakanath,
  Dwivedi, Gojayev, Gopalakrishnan, Gueudre, Hakkani-Tur, Hamza, H\"{u}ser,
  Jose, Khan, Liu, Lu, Manzotti, Natarajan, Owczarzak, Oz, Palumbo, Peris,
  Prakash, Rawls, Rosenbaum, Shenoy, Soltan, Sridhar, Tan, Triefenbach, Wei,
  Yu, Zheng, Tur, and Natarajan}]{FitzGerald2022AlexaTM}
Jack FitzGerald, Shankar Ananthakrishnan, Konstantine Arkoudas, Davide
  Bernardi, Abhishek Bhagia, Claudio Delli~Bovi, Jin Cao, Rakesh Chada, Amit
  Chauhan, Luoxin Chen, Anurag Dwarakanath, Satyam Dwivedi, Turan Gojayev,
  Karthik Gopalakrishnan, Thomas Gueudre, Dilek Hakkani-Tur, Wael Hamza,
  Jonathan~J. H\"{u}ser, Kevin~Martin Jose, Haidar Khan, Beiye Liu, Jianhua Lu,
  Alessandro Manzotti, Pradeep Natarajan, Karolina Owczarzak, Gokmen Oz, Enrico
  Palumbo, Charith Peris, Chandana~Satya Prakash, Stephen Rawls, Andy
  Rosenbaum, Anjali Shenoy, Saleh Soltan, Mukund~Harakere Sridhar, Lizhen Tan,
  Fabian Triefenbach, Pan Wei, Haiyang Yu, Shuai Zheng, Gokhan Tur, and Prem
  Natarajan. 2022.
\newblock \href {https://doi.org/10.1145/3534678.3539173} {Alexa teacher model:
  Pretraining and distilling multi-billion-parameter encoders for natural
  language understanding systems}.
\newblock In \emph{Proceedings of the 28th ACM SIGKDD Conference on Knowledge
  Discovery and Data Mining}, KDD '22, page 2893–2902, New York, NY, USA.
  Association for Computing Machinery.

\bibitem[{Hu et~al.(2020)Hu, Ruder, Siddhant, Neubig, Firat, and
  Johnson}]{pmlr-v119-hu20b}
Junjie Hu, Sebastian Ruder, Aditya Siddhant, Graham Neubig, Orhan Firat, and
  Melvin Johnson. 2020.
\newblock \href {https://proceedings.mlr.press/v119/hu20b.html} {{XTREME}: A
  massively multilingual multi-task benchmark for evaluating cross-lingual
  generalisation}.
\newblock In \emph{Proceedings of the 37th International Conference on Machine
  Learning}, volume 119 of \emph{Proceedings of Machine Learning Research},
  pages 4411--4421. PMLR.

\bibitem[{Kingma and Ba(2015)}]{Kingma2015AdamAM}
Diederik~P. Kingma and Jimmy Ba. 2015.
\newblock Adam: A method for stochastic optimization.
\newblock \emph{CoRR}, abs/1412.6980.

\bibitem[{Lewis et~al.(2020)Lewis, Liu, Goyal, Ghazvininejad, Mohamed, Levy,
  Stoyanov, and Zettlemoyer}]{Lewis2020BARTDS}
Mike Lewis, Yinhan Liu, Naman Goyal, Marjan Ghazvininejad, Abdelrahman Mohamed,
  Omer Levy, Veselin Stoyanov, and Luke Zettlemoyer. 2020.
\newblock Bart: Denoising sequence-to-sequence pre-training for natural
  language generation, translation, and comprehension.
\newblock In \emph{ACL}.

\bibitem[{Li et~al.(2021)Li, Arora, Chen, Gupta, Gupta, and
  Mehdad}]{li-etal-2021-mtop}
Haoran Li, Abhinav Arora, Shuohui Chen, Anchit Gupta, Sonal Gupta, and Yashar
  Mehdad. 2021.
\newblock \href {https://doi.org/10.18653/v1/2021.eacl-main.257} {{MTOP}: A
  comprehensive multilingual task-oriented semantic parsing benchmark}.
\newblock In \emph{Proceedings of the 16th Conference of the European Chapter
  of the Association for Computational Linguistics: Main Volume}, pages
  2950--2962, Online. Association for Computational Linguistics.

\bibitem[{Liu et~al.(2022)Liu, Huang, Lyu, Shakeri, Yu, and Li}]{liu2022enct5}
Frederick Liu, Terry Huang, Shihang Lyu, Siamak Shakeri, Hongkun Yu, and Jing
  Li. 2022.
\newblock \href {http://arxiv.org/abs/2110.08426} {Enct5: A framework for
  fine-tuning t5 as non-autoregressive models}.

\bibitem[{Liu et~al.(2020)Liu, Gu, Goyal, Li, Edunov, Ghazvininejad, Lewis, and
  Zettlemoyer}]{liu-etal-2020-multilingual-denoising}
Yinhan Liu, Jiatao Gu, Naman Goyal, Xian Li, Sergey Edunov, Marjan
  Ghazvininejad, Mike Lewis, and Luke Zettlemoyer. 2020.
\newblock \href {https://doi.org/10.1162/tacl_a_00343} {Multilingual denoising
  pre-training for neural machine translation}.
\newblock \emph{Transactions of the Association for Computational Linguistics},
  8:726--742.

\bibitem[{Liu et~al.(2019)Liu, Ott, Goyal, Du, Joshi, Chen, Levy, Lewis,
  Zettlemoyer, and Stoyanov}]{Liu2019RoBERTaAR}
Yinhan Liu, Myle Ott, Naman Goyal, Jingfei Du, Mandar Joshi, Danqi Chen, Omer
  Levy, Mike Lewis, Luke Zettlemoyer, and Veselin Stoyanov. 2019.
\newblock Roberta: A robustly optimized bert pretraining approach.
\newblock \emph{ArXiv}, abs/1907.11692.

\bibitem[{Nakayama(2018)}]{seqeval}
Hiroki Nakayama. 2018.
\newblock \href {https://github.com/chakki-works/seqeval} {{seqeval}: A python
  framework for sequence labeling evaluation}.
\newblock Software available from https://github.com/chakki-works/seqeval.

\bibitem[{Narayan et~al.(2018)Narayan, Cohen, and
  Lapata}]{narayan-etal-2018-dont}
Shashi Narayan, Shay~B. Cohen, and Mirella Lapata. 2018.
\newblock \href {https://doi.org/10.18653/v1/D18-1206} {Don{'}t give me the
  details, just the summary! topic-aware convolutional neural networks for
  extreme summarization}.
\newblock In \emph{Proceedings of the 2018 Conference on Empirical Methods in
  Natural Language Processing}, pages 1797--1807, Brussels, Belgium.
  Association for Computational Linguistics.

\bibitem[{Ni et~al.(2022)Ni, Hernandez~Abrego, Constant, Ma, Hall, Cer, and
  Yang}]{ni-etal-2022-sentence}
Jianmo Ni, Gustavo Hernandez~Abrego, Noah Constant, Ji~Ma, Keith Hall, Daniel
  Cer, and Yinfei Yang. 2022.
\newblock \href {https://doi.org/10.18653/v1/2022.findings-acl.146}
  {Sentence-t5: Scalable sentence encoders from pre-trained text-to-text
  models}.
\newblock In \emph{Findings of the Association for Computational Linguistics:
  ACL 2022}, pages 1864--1874, Dublin, Ireland. Association for Computational
  Linguistics.

\bibitem[{Nivre et~al.(2020)Nivre, de~Marneffe, Ginter, Hajivc, Manning,
  Pyysalo, Schuster, Tyers, and Zeman}]{Nivre2020UniversalDV}
Joakim Nivre, Marie-Catherine de~Marneffe, Filip Ginter, Jan Hajivc,
  Christopher~D. Manning, Sampo Pyysalo, Sebastian Schuster, Francis~M. Tyers,
  and Daniel Zeman. 2020.
\newblock Universal dependencies v2: An evergrowing multilingual treebank
  collection.
\newblock In \emph{International Conference on Language Resources and
  Evaluation}.

\bibitem[{Pan et~al.(2017)Pan, Zhang, May, Nothman, Knight, and
  Ji}]{pan-etal-2017-cross}
Xiaoman Pan, Boliang Zhang, Jonathan May, Joel Nothman, Kevin Knight, and Heng
  Ji. 2017.
\newblock \href {https://doi.org/10.18653/v1/P17-1178} {Cross-lingual name
  tagging and linking for 282 languages}.
\newblock In \emph{Proceedings of the 55th Annual Meeting of the Association
  for Computational Linguistics (Volume 1: Long Papers)}, pages 1946--1958,
  Vancouver, Canada. Association for Computational Linguistics.

\bibitem[{Radford and Narasimhan(2018)}]{Radford2018ImprovingLU}
Alec Radford and Karthik Narasimhan. 2018.
\newblock Improving language understanding by generative pre-training.

\bibitem[{Radford et~al.(2019)Radford, Wu, Child, Luan, Amodei, and
  Sutskever}]{Radford2019LanguageMA}
Alec Radford, Jeff Wu, Rewon Child, David Luan, Dario Amodei, and Ilya
  Sutskever. 2019.
\newblock Language models are unsupervised multitask learners.

\bibitem[{Raffel et~al.(2020{\natexlab{a}})Raffel, Shazeer, Roberts, Lee,
  Narang, Matena, Zhou, Li, and Liu}]{2020t5}
Colin Raffel, Noam Shazeer, Adam Roberts, Katherine Lee, Sharan Narang, Michael
  Matena, Yanqi Zhou, Wei Li, and Peter~J. Liu. 2020{\natexlab{a}}.
\newblock \href {http://jmlr.org/papers/v21/20-074.html} {Exploring the limits
  of transfer learning with a unified text-to-text transformer}.
\newblock \emph{Journal of Machine Learning Research}, 21(140):1--67.

\bibitem[{Raffel et~al.(2020{\natexlab{b}})Raffel, Shazeer, Roberts, Lee,
  Narang, Matena, Zhou, Li, and Liu}]{Raffel2020ExploringTL}
Colin Raffel, Noam~M. Shazeer, Adam Roberts, Katherine Lee, Sharan Narang,
  Michael Matena, Yanqi Zhou, Wei Li, and Peter~J. Liu. 2020{\natexlab{b}}.
\newblock Exploring the limits of transfer learning with a unified text-to-text
  transformer.
\newblock \emph{ArXiv}, abs/1910.10683.

\bibitem[{Rasley et~al.(2020)Rasley, Rajbhandari, Ruwase, and
  He}]{Rasley2020DeepSpeed}
Jeff Rasley, Samyam Rajbhandari, Olatunji Ruwase, and Yuxiong He. 2020.
\newblock \href {https://doi.org/10.1145/3394486.3406703} {Deepspeed: System
  optimizations enable training deep learning models with over 100 billion
  parameters}.
\newblock In \emph{Proceedings of the 26th ACM SIGKDD International Conference
  on Knowledge Discovery and Data Mining}, KDD '20, page 3505–3506, New York,
  NY, USA. Association for Computing Machinery.

\bibitem[{Rosenbaum et~al.(2022{\natexlab{a}})Rosenbaum, Soltan, Hamza,
  Damonte, Groves, and Saffari}]{rosenbaum-etal-2022-clasp}
Andy Rosenbaum, Saleh Soltan, Wael Hamza, Marco Damonte, Isabel Groves, and
  Amir Saffari. 2022{\natexlab{a}}.
\newblock \href {https://aclanthology.org/2022.aacl-short.56} {{CLASP}:
  Few-shot cross-lingual data augmentation for semantic parsing}.
\newblock In \emph{Proceedings of the 2nd Conference of the Asia-Pacific
  Chapter of the Association for Computational Linguistics and the 12th
  International Joint Conference on Natural Language Processing (Volume 2:
  Short Papers)}, pages 444--462, Online only. Association for Computational
  Linguistics.

\bibitem[{Rosenbaum et~al.(2022{\natexlab{b}})Rosenbaum, Soltan, Hamza,
  Versley, and Boese}]{rosenbaum-etal-2022-linguist}
Andy Rosenbaum, Saleh Soltan, Wael Hamza, Yannick Versley, and Markus Boese.
  2022{\natexlab{b}}.
\newblock \href {https://aclanthology.org/2022.coling-1.18} {{LINGUIST}:
  Language model instruction tuning to generate annotated utterances for intent
  classification and slot tagging}.
\newblock In \emph{Proceedings of the 29th International Conference on
  Computational Linguistics}, pages 218--241, Gyeongju, Republic of Korea.
  International Committee on Computational Linguistics.

\bibitem[{Rothe et~al.(2020)Rothe, Narayan, and
  Severyn}]{rothe-etal-2020-leveraging}
Sascha Rothe, Shashi Narayan, and Aliaksei Severyn. 2020.
\newblock \href {https://doi.org/10.1162/tacl_a_00313} {Leveraging pre-trained
  checkpoints for sequence generation tasks}.
\newblock \emph{Transactions of the Association for Computational Linguistics},
  8:264--280.

\bibitem[{Sanh et~al.(2021)Sanh, Webson, Raffel, Bach, Sutawika, Alyafeai,
  Chaffin, Stiegler, Scao, Raja, Dey, BARI, Xu, Thakker, Sharma, Szczechla,
  Kim, Chhablani, Nayak, Datta, Chang, Jiang, Wang, Manica, Shen, Yong, Pandey,
  Bawden, Wang, Neeraj, Rozen, Sharma, Santilli, F{\'e}vry, Fries, Teehan,
  Biderman, Gao, Bers, Wolf, and Rush}]{Sanh2021MultitaskPT}
Victor Sanh, Albert Webson, Colin Raffel, Stephen~H. Bach, Lintang~A. Sutawika,
  Zaid Alyafeai, Antoine Chaffin, Arnaud Stiegler, Teven~Le Scao, Arun Raja,
  Manan Dey, M~SAIFUL BARI, Canwen Xu, Urmish Thakker, Shanya Sharma, Eliza
  Szczechla, Taewoon Kim, Gunjan Chhablani, Nihal~V. Nayak, Debajyoti Datta,
  Jonathan Chang, Mike Tian-Jian Jiang, Han Wang, Matteo Manica, Sheng Shen,
  Zheng~Xin Yong, Harshit Pandey, Rachel Bawden, Thomas Wang, Trishala Neeraj,
  Jos Rozen, Abheesht Sharma, Andrea Santilli, Thibault F{\'e}vry, Jason~Alan
  Fries, Ryan Teehan, Stella~Rose Biderman, Leo Gao, T.~G.~Owe Bers, Thomas
  Wolf, and Alexander~M. Rush. 2021.
\newblock Multitask prompted training enables zero-shot task generalization.
\newblock \emph{ArXiv}, abs/2110.08207.

\bibitem[{Soltan et~al.(2022)Soltan, Ananthakrishnan, FitzGerald, Gupta, Hamza,
  Khan, Peris, Rawls, Rosenbaum, Rumshisky, Prakash, Sridhar, Triefenbach,
  Verma, Tur, and Natarajan}]{Soltan2022AlexaTM2F}
Saleh Soltan, Shankar Ananthakrishnan, Jack G.~M. FitzGerald, Rahul Gupta, Wael
  Hamza, Haidar Khan, Charith~S. Peris, Stephen Rawls, Andrew Rosenbaum, Anna
  Rumshisky, Chandan Prakash, Mukund Sridhar, Fabian Triefenbach, Apurv Verma,
  Gokhan Tur, and Premkumar Natarajan. 2022.
\newblock Alexatm 20b: Few-shot learning using a large-scale multilingual
  seq2seq model.
\newblock \emph{ArXiv}, abs/2208.01448.

\bibitem[{Thoppilan et~al.(2022)Thoppilan, Freitas, Hall, Shazeer,
  Kulshreshtha, Cheng, Jin, Bos, Baker, Du, Li, Lee, Zheng, Ghafouri, Menegali,
  Huang, Krikun, Lepikhin, Qin, Chen, Xu, Chen, Roberts, Bosma, Zhou, Chang,
  Krivokon, Rusch, Pickett, Meier-Hellstern, Morris, Doshi, Santos, Duke,
  S{\o}raker, Zevenbergen, Prabhakaran, D{\'i}az, Hutchinson, Olson, Molina,
  Hoffman-John, Lee, Aroyo, Rajakumar, Butryna, Lamm, Kuzmina, Fenton, Cohen,
  Bernstein, Kurzweil, Aguera-Arcas, Cui, Croak, Chi, and
  Le}]{Thoppilan2022LaMDALM}
Romal Thoppilan, Daniel~De Freitas, Jamie Hall, Noam~M. Shazeer, Apoorv
  Kulshreshtha, Heng-Tze Cheng, Alicia Jin, Taylor Bos, Leslie Baker, Yu~Du,
  Yaguang Li, Hongrae Lee, Huaixiu Zheng, Amin Ghafouri, Marcelo Menegali,
  Yanping Huang, Maxim Krikun, Dmitry Lepikhin, James Qin, Dehao Chen,
  Yuanzhong Xu, Zhifeng Chen, Adam Roberts, Maarten Bosma, Yanqi Zhou,
  Chung-Ching Chang, I.~A. Krivokon, Willard~James Rusch, Marc Pickett,
  Kathleen~S. Meier-Hellstern, Meredith~Ringel Morris, Tulsee Doshi,
  Renelito~Delos Santos, Toju Duke, Johnny~Hartz S{\o}raker, Ben Zevenbergen,
  Vinodkumar Prabhakaran, Mark D{\'i}az, Ben Hutchinson, Kristen Olson,
  Alejandra Molina, Erin Hoffman-John, Josh Lee, Lora Aroyo, Ravindran
  Rajakumar, Alena Butryna, Matthew Lamm, V.~O. Kuzmina, Joseph Fenton, Aaron
  Cohen, Rachel Bernstein, Ray Kurzweil, Blaise Aguera-Arcas, Claire Cui,
  Marian Croak, Ed~Chi, and Quoc Le. 2022.
\newblock Lamda: Language models for dialog applications.
\newblock \emph{ArXiv}, abs/2201.08239.

\bibitem[{Vaswani et~al.(2017)Vaswani, Shazeer, Parmar, Uszkoreit, Jones,
  Gomez, Kaiser, and Polosukhin}]{Vaswani2017AttentionIA}
Ashish Vaswani, Noam~M. Shazeer, Niki Parmar, Jakob Uszkoreit, Llion Jones,
  Aidan~N. Gomez, Lukasz Kaiser, and Illia Polosukhin. 2017.
\newblock Attention is all you need.
\newblock \emph{ArXiv}, abs/1706.03762.

\bibitem[{Wolf et~al.(2020)Wolf, Debut, Sanh, Chaumond, Delangue, Moi, Cistac,
  Rault, Louf, Funtowicz, Davison, Shleifer, von Platen, Ma, Jernite, Plu, Xu,
  Le~Scao, Gugger, Drame, Lhoest, and Rush}]{wolf-etal-2020-transformers}
Thomas Wolf, Lysandre Debut, Victor Sanh, Julien Chaumond, Clement Delangue,
  Anthony Moi, Pierric Cistac, Tim Rault, Remi Louf, Morgan Funtowicz, Joe
  Davison, Sam Shleifer, Patrick von Platen, Clara Ma, Yacine Jernite, Julien
  Plu, Canwen Xu, Teven Le~Scao, Sylvain Gugger, Mariama Drame, Quentin Lhoest,
  and Alexander Rush. 2020.
\newblock \href {https://doi.org/10.18653/v1/2020.emnlp-demos.6} {Transformers:
  State-of-the-art natural language processing}.
\newblock In \emph{Proceedings of the 2020 Conference on Empirical Methods in
  Natural Language Processing: System Demonstrations}, pages 38--45, Online.
  Association for Computational Linguistics.

\bibitem[{Xiong et~al.(2020)Xiong, Yang, He, Zheng, Zheng, Xing, Zhang, Lan,
  Wang, and Liu}]{Xiong2020OnLN}
Ruibin Xiong, Yunchang Yang, Di~He, Kai Zheng, Shuxin Zheng, Chen Xing,
  Huishuai Zhang, Yanyan Lan, Liwei Wang, and Tie-Yan Liu. 2020.
\newblock On layer normalization in the transformer architecture.
\newblock \emph{ArXiv}, abs/2002.04745.

\bibitem[{Xu et~al.(2020)Xu, Haider, and Mansour}]{xu-etal-2020-end}
Weijia Xu, Batool Haider, and Saab Mansour. 2020.
\newblock \href {https://doi.org/10.18653/v1/2020.emnlp-main.410} {End-to-end
  slot alignment and recognition for cross-lingual {NLU}}.
\newblock In \emph{Proceedings of the 2020 Conference on Empirical Methods in
  Natural Language Processing (EMNLP)}, pages 5052--5063, Online. Association
  for Computational Linguistics.

\bibitem[{Xue et~al.(2021)Xue, Constant, Roberts, Kale, Al-Rfou, Siddhant,
  Barua, and Raffel}]{xue-etal-2021-mt5}
Linting Xue, Noah Constant, Adam Roberts, Mihir Kale, Rami Al-Rfou, Aditya
  Siddhant, Aditya Barua, and Colin Raffel. 2021.
\newblock \href {https://doi.org/10.18653/v1/2021.naacl-main.41} {m{T}5: A
  massively multilingual pre-trained text-to-text transformer}.
\newblock In \emph{Proceedings of the 2021 Conference of the North American
  Chapter of the Association for Computational Linguistics: Human Language
  Technologies}, pages 483--498, Online. Association for Computational
  Linguistics.

\bibitem[{Zhang et~al.(2022)Zhang, Roller, Goyal, Artetxe, Chen, Chen, Dewan,
  Diab, Li, Lin, Mihaylov, Ott, Shleifer, Shuster, Simig, Koura, Sridhar, Wang,
  and Zettlemoyer}]{Zhang2022OPTOP}
Susan Zhang, Stephen Roller, Naman Goyal, Mikel Artetxe, Moya Chen, Shuohui
  Chen, Christopher Dewan, Mona Diab, Xian Li, Xi~Victoria Lin, Todor Mihaylov,
  Myle Ott, Sam Shleifer, Kurt Shuster, Daniel Simig, Punit~Singh Koura, Anjali
  Sridhar, Tianlu Wang, and Luke Zettlemoyer. 2022.
\newblock Opt: Open pre-trained transformer language models.
\newblock \emph{ArXiv}, abs/2205.01068.

\end{thebibliography}
\bibliographystyle{acl_natbib}

\onecolumn
\appendix

\section{Additional Related Work}
\label{sec:additional_related_work}

Pre-trained Transformer models \citep{Vaswani2017AttentionIA} are commonly used in Natural Language Processing (NLP) for both transfer learning in downstream tasks \citep{devlin-etal-2019-bert,Liu2019RoBERTaAR,Radford2018ImprovingLU,Radford2019LanguageMA} and for in-context learning \citep{Brown2020LanguageMA}. Transformers were originally designed as sequence-to-sequence (seq2seq) models with an encoder and a decoder component \citep{Vaswani2017AttentionIA}. However, all three obvious variants of this architecture are now common:
encoder-only \citep{devlin-etal-2019-bert}, decoder-only \citep{Radford2018ImprovingLU,Radford2019LanguageMA,Brown2020LanguageMA,Chowdhery2022PaLMSL,Zhang2022OPTOP,Thoppilan2022LaMDALM} and seq2seq \citep{Lewis2020BARTDS,Raffel2020ExploringTL,Sanh2021MultitaskPT,Dong2019UnifiedLM,Bao2020UniLMv2PL}.  

Commonly, encoder transformer models are pre-trained using the MLM objective \citep{devlin-etal-2019-bert}. Decoders are pre-trained using a next-token left-to-right prediction (causal) language modeling objective \citep{Radford2018ImprovingLU} or some version of autoregressive de-noising \citep{Lewis2020BARTDS}. Seq2seq models often combine these objectives \citep{Lewis2020BARTDS,Bao2020UniLMv2PL}.

We follow the multilingual approach of models such as XLM-RoBERTa \citep{conneau-etal-2020-unsupervised} (encoder-only) and
mT5/mBART \citep{xue-etal-2021-mt5,liu-etal-2020-multilingual-denoising} (seq2seq),
where the model is pre-trained on data from multiple languages.
This enables \textbf{cross-lingual zero-shot fine-tuning}, where the model is fine-tuned on task
data only from a single language (usually English), then evaluated on multiple languages.

Previous literature has explored using a pre-trained encoder to initialize a larger encoder \citep{chen-etal-2022-bert2bert}
or a seq2seq model \citep{rothe-etal-2020-leveraging}.
The latter was applied to large-scale models, e.g. \mbox{AlexaTM 20B} and \mbox{AlexaTM 5B} \citep{Soltan2022AlexaTM2F,rosenbaum-etal-2022-linguist,FitzGerald2022AlexaTM}.
Our work provides the first direct comparison of warm-starting vs. from-scratch seq2seq pre-training
using the same data and codebase.

Recently, Sentence-T5 \citep{ni-etal-2022-sentence} studied the opposite direction,
showing that extracting the encoder from
T5 \citep{2020t5} can out-perform BERT on several sentence-level tasks.
We also explore extracting the encoder from a seq2seq model,
adding the novelty of the first explicit comparison
with MLM encoders
using the same pre-training data and codebase.
Furthermore, whereas Sentence-T5 studies only sentence level tasks in English, we study both
sentence-level and token-level (e.g. sequence labeling) multilingual tasks.
We show that the encoder extracted from a seq2seq model under-performs
on token-level tasks, motivating our proposed sequential pre-training recipes.

EncT5 \citep{liu2022enct5} proposes an alternative method to fine-tune the encoder from a seq2seq model
for classification and sequence labeling tasks, by attaching a randomly initialized one-layer decoder with cross-attention.
They report substantial improvements on UDPOS (Part-of-Speech tagging, a sequence labeling task)
compared to an mBERT (MLM encoder) model of similar encoder size,
however the comparison is between models pre-trained on different data and codebases.
For a cleaner comparison, we would need to implement and evaluate the EncT5
framework with our models, which is challenging since no reference implementation is available, and also because \citet{liu2022enct5}
provide only the average number across languages for UDPOS and
do not report per language.
Therefore, we defer a more thorough study of EncT5 vs. standard feed-forward layer classification heads to future work.

\section{Results by Language}
\label{sec:res_all_langs}

Our main results in Section \ref{sec:ft_res} (Tables \ref{table:enc_all} and \ref{table:generative})
show only the English and average zero-shot results for brevity.
Here, for completeness, we
show the results on each langauge for
XNLI (Table \ref{table:xnli_full}),
mATIS++ Intent Classification (IC) (Table \ref{table:matispp_ic_full}),
mATIS++ Slot Labeling (SL) (Table \ref{table:matispp_sl_full}),
WikiANN NER (Table \ref{table:wikiann_full}),
UDPOS (Table \ref{table:udpos_full}),
and
mTOP semantic parsing (Table \ref{table:mtop_full}).

\begin{table*}[h]
\footnotesize
\begin{center}
\begin{tabular}{lc|ccccc | c}
\toprule
Encoder & en & ar & de & es & fr & hi & avg-0s\\ 
\midrule
\multicolumn{8}{c}{Encoder Model From Scratch (MLM only)} \\
\midrule
  roberta-12e       & $\textbf{84.5}_{\pm{}0.5}$ & $\textbf{72.9}_{\pm{}0.2}$ & $\textbf{76.9}_{\pm{}0.3}$ & $\textbf{79.9}_{\pm{}0.2}$ & $\textbf{78.7}_{\pm{}0.3}$ & $\textbf{70.5}_{\pm{}0.8}$ & $\textbf{75.8}_{\pm{}0.2}$ \\
  \midrule
\multicolumn{8}{c}{Encoder of Seq2Seq Models (de-noising only)} \\
 \midrule
 bart-12e12d       & $\underline{83.9}_{\pm{}0.2}$ & $71.6_{\pm{}0.6}$ & $76.0_{\pm{}0.5}$ & $79.2_{\pm{}0.8}$ & $77.8_{\pm{}0.1}$ & $\underline{68.9}_{\pm{}0.3}$ & $74.7_{\pm{}0.3}$ \\
 bart-12e12d-mask & $\underline{83.9}_{\pm{}0.4}$ & $\underline{71.9}_{\pm{}0.7}$ & $\underline{76.3}_{\pm{}0.2}$ & $\underline{79.5}_{\pm{}0.6}$ & $\underline{78.5}_{\pm{}0.5}$ & $68.5_{\pm{}1.3}$ & $\underline{75.0}_{\pm{}0.6}$ \\
 \midrule
 bart-12e2d        & $71.3_{\pm{}0.1}$ & $56.7_{\pm{}0.7}$ & $60.3_{\pm{}0.3}$ & $64.2_{\pm{}0.7}$ & $63.9_{\pm{}0.6}$ & $53.3_{\pm{}0.2}$ & $59.7_{\pm{}0.5}$ \\
 bart-12e2d-mask   & $82.9_{\pm{}0.3}$ & $70.9_{\pm{}0.4}$ & $74.7_{\pm{}0.5}$ & $78.1_{\pm{}0.3}$ & $76.9_{\pm{}0.4}$ & $68.2_{\pm{}0.5}$ & $73.8_{\pm{}0.2}$ \\
 bart-12e1d-mask   & $82.4_{\pm{}0.2}$ & $69.6_{\pm{}0.3}$ & $73.5_{\pm{}0.3}$ & $77.0_{\pm{}0.1}$ & $76.3_{\pm{}0.4}$ & $66.9_{\pm{}0.2}$ & $72.7_{\pm{}0.1}$ \\
\midrule
\multicolumn{8}{c}{Recipe 1: Encoder of Seq2Seq Model + MLM} \\
\midrule
  bart-12e12d+mlm  & $80.3_{\pm{}0.4}$ & $65.6_{\pm{}0.4}$ & $70.6_{\pm{}0.2}$ & $72.9_{\pm{}0.8}$ & $72.6_{\pm{}0.2}$ & $63.4_{\pm{}0.8}$ & $69.0_{\pm{}0.4}$ \\
\bottomrule
\end{tabular}
\end{center}
\caption{Encoder model results by language on XNLI test sets: accuracy.}
\label{table:xnli_full}
\end{table*}

\begin{table*}[h]
\footnotesize
\begin{center}
\begin{tabular}{lc|cccccc | c}
\toprule
Encoder & en & de & es & fr & hi & ja & pt & avg-0s \\
\midrule
\multicolumn{9}{c}{Encoder Model From Scratch (MLM only)} \\
\midrule
   roberta-12e      & $\textbf{97.8}_{\pm{}0.1}$ & $\textbf{92.7}_{\pm{}2.2}$ & $\textbf{96.2}_{\pm{}0.5}$ & $\underline{94.6}_{\pm{}1.4}$ & $\textbf{79.5}_{\pm{}4.5}$ & $65.6_{\pm{}17.1}$ & $\underline{94.3}_{\pm{}2.4}$ & $87.2_{\pm{}4.1}$ \\
  \midrule
\multicolumn{9}{c}{Encoder of Seq2Seq Models (de-noising only)} \\
  \midrule
 bart-12e12d       & $96.8_{\pm{}0.1}$ & $91.0_{\pm{}2.5}$ & $91.0_{\pm{}0.4}$ & $93.1_{\pm{}1.5}$ & $77.7_{\pm{}3.6}$ & $72.1_{\pm{}4.5}$ & $92.2_{\pm{}1.8}$ & $86.2_{\pm{}1.5}$ \\
 bart-12e12d-mask & $97.1_{\pm{}0.1}$ & $89.7_{\pm{}1.1}$ & $94.2_{\pm{}0.4}$ & $94.0_{\pm{}0.8}$ & $78.6_{\pm{}0.7}$ & $\underline{75.0}_{\pm{}3.4}$ & $91.9_{\pm{}1.1}$ & $87.3_{\pm{}0.7}$ \\
 \midrule
 bart-12e2d        & $96.1_{\pm{}0.1}$ & $80.4_{\pm{}8.1}$ & $84.7_{\pm{}3.5}$ & $86.1_{\pm{}3.0}$ & $74.3_{\pm{}1.3}$ & $64.4_{\pm{}5.6}$ & $84.6_{\pm{}2.6}$ & $79.1_{\pm{}0.8}$ \\
 bart-12e2d-mask   & $96.8_{\pm{}0.1}$ & $\underline{92.4}_{\pm{}0.6}$ & $94.5_{\pm{}0.5}$ & $\textbf{94.7}_{\pm{}0.5}$ & $79.1_{\pm{}1.5}$ & $73.9_{\pm{}5.1}$ & $\textbf{94.4}_{\pm{}0.4}$ & $\textbf{88.1}_{\pm{}0.9}$ \\
 bart-12e1d-mask   & $97.0_{\pm{}0.1}$ & $90.1_{\pm{}0.8}$ & $\underline{94.8}_{\pm{}0.4}$ & $93.3_{\pm{}0.4}$ & $\underline{79.4}_{\pm{}0.8}$ & $\textbf{76.5}_{\pm{}3.0}$ & $91.6_{\pm{}0.4}$ & $\underline{87.6}_{\pm{}0.5}$ \\
\midrule
\multicolumn{9}{c}{Recipe 1: Encoder of Seq2Seq Model + MLM} \\
\midrule
 bart-12e12d+mlm  & $\underline{97.2}_{\pm{}0.4}$ & $86.1_{\pm{}4.2}$ & $92.0_{\pm{}0.8}$ & $91.1_{\pm{}1.5}$ & $76.1_{\pm{}2.7}$ & $70.1_{\pm{}4.4}$ & $88.2_{\pm{}3.3}$ & $83.9_{\pm{}1.6}$ \\
\bottomrule
\end{tabular}
\end{center}
\caption{Encoder model results by language on mATIS++ test sets, Intent Classificaiton (IC) accuracy.}
\label{table:matispp_ic_full}
\end{table*}

\begin{table*}[h]
\footnotesize
\begin{center}
\begin{tabular}{lc|cccccc | c}
\toprule
Encoder & en & de & es & fr & hi & ja & pt & avg-0s \\
\midrule
\multicolumn{9}{c}{Encoder Model From Scratch (MLM only)} \\
\midrule
   roberta-12e      & $\textbf{95.7}_{\pm{}0.1}$ & $\textbf{82.8}_{\pm{}1.2}$ & $\textbf{81.8}_{\pm{}0.6}$ & $\textbf{72.3}_{\pm{}1.7}$ & $\underline{31.8}_{\pm{}1.3}$ & $\textbf{20.9}_{\pm{}1.2}$ & $\textbf{79.8}_{\pm{}0.8}$ & $\textbf{61.6}_{\pm{}0.6}$ \\
  \midrule
\multicolumn{9}{c}{Encoder of Seq2Seq Models (de-noising only)} \\
  \midrule
 bart-12e12d       & $92.5_{\pm{}0.3}$ & $57.0_{\pm{}4.8}$ & $52.6_{\pm{}0.9}$ & $58.5_{\pm{}0.4}$ & $25.1_{\pm{}1.3}$ & $12.6_{\pm{}1.5}$ & $60.0_{\pm{}3.1}$ & $44.3_{\pm{}1.3}$ \\
 bart-12e12d-mask & $91.1_{\pm{}0.9}$ & $54.7_{\pm{}2.6}$ & $52.9_{\pm{}2.4}$ & $52.5_{\pm{}0.9}$ & $23.2_{\pm{}2.9}$ & $10.5_{\pm{}1.5}$ & $53.9_{\pm{}1.7}$ & $41.3_{\pm{}1.3}$ \\
 \midrule
 bart-12e2d        & $91.4_{\pm{}0.1}$ & $52.0_{\pm{}4.1}$ & $53.2_{\pm{}2.3}$ & $50.3_{\pm{}1.7}$ & $11.4_{\pm{}0.8}$ & $3.9_{\pm{}1.5}$  & $58.3_{\pm{}2.2}$ & $38.2_{\pm{}1.7}$ \\
 bart-12e2d-mask   & $92.3_{\pm{}0.3}$ & $63.7_{\pm{}2.5}$ & $60.7_{\pm{}1.1}$ & $60.8_{\pm{}1.8}$ & $26.0_{\pm{}1.5}$ & $10.4_{\pm{}0.9}$ & $66.4_{\pm{}2.1}$ & $48.0_{\pm{}1.4}$ \\
 bart-12e1d-mask   & $92.8_{\pm{}0.5}$ & $65.6_{\pm{}4.8}$ & $59.5_{\pm{}0.4}$ & $61.7_{\pm{}0.3}$ & $24.7_{\pm{}1.7}$ & $\underline{16.3}_{\pm{}1.8}$ & $68.0_{\pm{}1.6}$ & $49.3_{\pm{}1.2}$ \\
\midrule
\multicolumn{9}{c}{Recipe 1: Encoder of Seq2Seq Model + MLM} \\
\midrule
 bart-12e12d+mlm  & $\underline{95.3}_{\pm{}0.2}$ & $\underline{70.5}_{\pm{}7.6}$ & $\underline{79.7}_{\pm{}1.2}$ & $\underline{67.3}_{\pm{}1.4}$ & $\textbf{33.5}_{\pm{}3.9}$ & $15.9_{\pm{}3.8}$ & $\underline{72.2}_{\pm{}0.6}$ & $\underline{56.5}_{\pm{}2.8}$ \\
\bottomrule
\end{tabular}
\end{center}
\caption{Encoder model results by language on mATIS++ test sets, Slot Labeling (SL) f1 score.}
\label{table:matispp_sl_full}
\end{table*}

\begin{table*}[h]
\footnotesize
\setlength{\tabcolsep}{4pt}
\begin{center}
\begin{tabular}{lc|ccccccccccc | c}
\toprule
 Encoder            &   en &   ar &   de &   es &   fr &   hi &   it &   ja &   mr &   pt &   ta &   te &   avg-0s \\
\midrule
 \multicolumn{14}{c}{Encoder Model From Scratch (MLM only)} \\
 \midrule
 \multirow{2}{*}{roberta-12e}      & \textbf{83.0}                     & \textbf{45.7}                     & \textbf{73.5}                     & \textbf{68.9}                     & \textbf{75.4}                     & \textbf{71.5}                     & \textbf{76.7}                     & \textbf{28.0}                     & \textbf{57.5}                     & \textbf{74.7}                     & \textbf{53.9}                     & \textbf{46.5}                     & \textbf{61.1}                     \\
                                   & $\pm{}0.1$ & $\pm{}2.1$ & $\pm{}0.7$ & $\pm{}0.5$ & $\pm{}0.4$ & $\pm{}0.5$ & $\pm{}0.7$ & $\pm{}1.0$ & $\pm{}2.0$ & $\pm{}0.2$ & $\pm{}0.3$ & $\pm{}0.7$ & $\pm{}0.4$ \\
\midrule
   \multicolumn{14}{c}{Encoder of Seq2Seq Models (de-noising only)} \\
 \midrule
 \multirow{2}{*}{bart-12e12d}      & 76.6                     & 44.4                     & \underline{64.8}                     & 61.1                     & 70.7                     & 62.2                     & 69.7                     & 10.2                     & 41.0                     & 69.7                     & 42.0                     & 37.0                     & 52.1                     \\
                                   & $\pm{}0.2$ & $\pm{}1.8$ & $\pm{}1.2$ & $\pm{}1.7$ & $\pm{}0.8$ & $\pm{}2.2$ & $\pm{}0.7$ & $\pm{}0.6$ & $\pm{}1.9$ & $\pm{}0.4$ & $\pm{}0.2$ & $\pm{}2.3$ & $\pm{}0.9$ \\
 \midrule
 \multirow{2}{*}{bart-12e12d-mask} & 73.2                     & 30.6                     & 57.7                     & 59.8                     & 67.4                     & 60.7                     & 66.1                     & 8.1                      & 41.3                     & 68.9                     & 37.6                     & 33.8                     & 48.4                     \\
                                   & $\pm{}0.1$ & $\pm{}0.8$ & $\pm{}0.3$ & $\pm{}0.5$ & $\pm{}0.5$ & $\pm{}0.1$ & $\pm{}0.5$ & $\pm{}0.7$ & $\pm{}2.9$ & $\pm{}1.5$ & $\pm{}1.1$ & $\pm{}1.9$ & $\pm{}0.6$ \\

\midrule
 \multirow{2}{*}{bart-12e2d}       & 69.3                     & 31.0                     & 53.5                     & 54.9                     & 61.2                     & 48.7                     & 62.0                     & 7.1                      & 33.2                     & 61.9                     & 31.1                     & 28.0                     & 42.9                     \\
                                   & $\pm{}0.5$ & $\pm{}1.7$ & $\pm{}1.0$ & $\pm{}0.5$ & $\pm{}0.5$ & $\pm{}0.8$ & $\pm{}0.8$ & $\pm{}0.4$ & $\pm{}3.3$ & $\pm{}0.2$ & $\pm{}0.8$ & $\pm{}1.0$ & $\pm{}0.1$ \\
 \midrule
 \multirow{2}{*}{bart-12e2d-mask}  & 76.5                     & \underline{45.2}                     & 63.5                     & \underline{65.0}                     & \underline{70.8}                     & \underline{64.0}                     & \underline{69.9}                     & 10.3                     & \underline{44.4}                     & \underline{71.5}                     & \underline{47.7}                     & \underline{41.4}                     & \underline{54.0}                     \\
                                   & $\pm{}0.2$ & $\pm{}1.9$ & $\pm{}0.5$ & $\pm{}2.4$ & $\pm{}0.5$ & $\pm{}1.0$ & $\pm{}0.3$ & $\pm{}0.7$ & $\pm{}3.3$ & $\pm{}0.5$ & $\pm{}2.4$ & $\pm{}3.1$ & $\pm{}0.6$ \\
 \midrule
 \multirow{2}{*}{bart-12e1d-mask}  & 74.6                     & 40.9                     & 54.5                     & 64.0                     & 64.3                     & 54.3                     & 65.4                     & 9.4                      & 43.0                     & 67.3                     & 37.8                     & 32.1                     & 48.5                     \\
                                   & $\pm{}0.5$ & $\pm{}2.5$ & $\pm{}1.6$ & $\pm{}2.0$ & $\pm{}0.6$ & $\pm{}1.0$ & $\pm{}0.3$ & $\pm{}0.8$ & $\pm{}0.4$ & $\pm{}0.5$ & $\pm{}0.9$ & $\pm{}2.3$ & $\pm{}0.3$ \\
 \midrule
\multicolumn{14}{c}{Recipe 1: Encoder of Seq2Seq Model + MLM} \\
\midrule
 \multirow{2}{*}{bart-12e12d+mlm}  & \underline{79.9}                     & 29.8                     & 62.8                     & 60.9                     & 68.9                     & 58.7                     & 68.7                     & \underline{13.6}                     & 29.4                     & 69.5                     & 33.7                     & 27.0                     & 47.5                     \\
                                   & $\pm{}0.2$ & $\pm{}1.0$ & $\pm{}0.6$ & $\pm{}0.5$ & $\pm{}0.3$ & $\pm{}1.4$ & $\pm{}0.4$ & $\pm{}0.3$ & $\pm{}0.8$ & $\pm{}0.7$ & $\pm{}1.1$ & $\pm{}0.9$ & $\pm{}0.5$ \\
\bottomrule
\end{tabular}
\end{center}
\caption{Encoder model results by language on WikiANN Named Entity Recognition (NER) test sets, f1 score.}
\label{table:wikiann_full}
\end{table*}

\begin{table*}[h]
\footnotesize
\setlength{\tabcolsep}{4pt}
\begin{center}
\begin{tabular}{lc|ccccccccccc | c}
\toprule
 Encoder            &   en &   ar &   de &   es &   fr &   hi &   it &   ja &   mr &   pt &   ta &   te &   avg-0s \\
\midrule
 \multicolumn{14}{c}{Encoder Model From Scratch (MLM only)} \\
 \midrule
  \multirow{2}{*}{roberta-12e}      & \textbf{95.8}                     & \textbf{65.9}                     & \textbf{86.2}                     & \textbf{85.6}                     & \textbf{77.6}                     & \textbf{67.5}                     & \textbf{88.0}                     & \textbf{44.9}                     & \textbf{74.4}                     & \textbf{86.1}                     & \textbf{55.1}                     & \textbf{76.8}                     & \textbf{73.5}                     \\
                                   & $\pm{}0.0$ & $\pm{}0.2$ & $\pm{}0.2$ & $\pm{}0.5$ & $\pm{}0.2$ & $\pm{}0.6$ & $\pm{}0.4$ & $\pm{}1.3$ & $\pm{}1.7$ & $\pm{}0.3$ & $\pm{}0.1$ & $\pm{}1.2$ & $\pm{}0.2$ \\
 \midrule
   \multicolumn{14}{c}{Encoder of Seq2Seq Models (de-noising only)} \\
 \midrule
 \multirow{2}{*}{bart-12e12d}      & 94.3                     & \underline{54.0}                     & 75.0                     & 70.7                     & 66.7                     & \underline{57.0}                     & 71.2                     & 32.1                     & 61.5                     & 72.8                     & 47.4                     & 68.6                     & \underline{61.5}                     \\
                                   & $\pm{}0.7$ & $\pm{}0.4$ & $\pm{}3.9$ & $\pm{}5.5$ & $\pm{}1.1$ & $\pm{}1.4$ & $\pm{}2.4$ & $\pm{}7.9$ & $\pm{}9.7$ & $\pm{}3.0$ & $\pm{}4.5$ & $\pm{}7.8$ & $\pm{}0.4$ \\
 \midrule
 \multirow{2}{*}{bart-12e12d-mask} & 93.3                     & 49.0                     & 62.4                     & 58.9                     & 56.4                     & 50.6                     & 60.9                     & 27.3                     & 60.0                     & 64.3                     & 47.3                     & 69.0                     & 55.1                     \\
                                   & $\pm{}0.1$ & $\pm{}1.3$ & $\pm{}1.2$ & $\pm{}0.5$ & $\pm{}0.4$ & $\pm{}1.4$ & $\pm{}0.6$ & $\pm{}1.1$ & $\pm{}1.1$ & $\pm{}0.5$ & $\pm{}0.5$ & $\pm{}0.9$ & $\pm{}0.4$ \\
 \midrule
 \multirow{2}{*}{bart-12e2d}       & 92.1                     & 43.5                     & 60.8                     & 58.4                     & 54.6                     & 42.3                     & 58.5                     & 16.8                     & 57.2                     & 63.2                     & 41.4                     & 61.1                     & 50.7                     \\
                                   & $\pm{}0.1$ & $\pm{}0.9$ & $\pm{}1.5$ & $\pm{}1.7$ & $\pm{}1.4$ & $\pm{}0.3$ & $\pm{}1.3$ & $\pm{}0.3$ & $\pm{}3.0$ & $\pm{}0.9$ & $\pm{}0.3$ & $\pm{}1.2$ & $\pm{}0.5$ \\
 \midrule
 \multirow{2}{*}{bart-12e2d-mask}  & 93.3                     & 48.9                     & 61.7                     & 52.8                     & 52.9                     & 48.8                     & 58.1                     & 27.1                     & \underline{63.3}                     & 59.3                     & \underline{48.1}                     & \underline{73.4}                     & 54.0                     \\
                                   & $\pm{}0.1$ & $\pm{}0.6$ & $\pm{}2.6$ & $\pm{}0.9$ & $\pm{}1.8$ & $\pm{}0.5$ & $\pm{}1.2$ & $\pm{}1.2$ & $\pm{}1.5$ & $\pm{}1.6$ & $\pm{}1.0$ & $\pm{}2.2$ & $\pm{}0.6$ \\
 \midrule
 \multirow{2}{*}{bart-12e1d-mask}  & 92.4                     & 44.8                     & 52.5                     & 43.8                     & 43.8                     & 43.0                     & 47.9                     & 19.8                     & 58.4                     & 53.0                     & 42.1                     & 60.0                     & 46.3                     \\
                                   & $\pm{}0.1$ & $\pm{}1.4$ & $\pm{}3.4$ & $\pm{}1.6$ & $\pm{}2.8$ & $\pm{}2.6$ & $\pm{}3.2$ & $\pm{}1.7$ & $\pm{}1.9$ & $\pm{}1.5$ & $\pm{}1.1$ & $\pm{}1.9$ & $\pm{}1.7$ \\
  \midrule
\multicolumn{14}{c}{Recipe 1: Encoder of Seq2Seq Model + MLM} \\
\midrule
 \multirow{2}{*}{bart-12e12d+mlm}  & \underline{95.1}                     & 53.5                     & \underline{78.2}                     & \underline{76.1}                     & \underline{68.2}                     & 56.0                     & \underline{72.8}                     & \underline{39.6}                     & 49.4                     & \underline{74.9}                     & 41.9                     & 57.5                     & 60.7                     \\
                                   & $\pm{}0.0$ & $\pm{}1.3$ & $\pm{}1.1$ & $\pm{}0.7$ & $\pm{}1.8$ & $\pm{}2.0$ & $\pm{}0.7$ & $\pm{}1.4$ & $\pm{}1.0$ & $\pm{}1.0$ & $\pm{}0.3$ & $\pm{}1.9$ & $\pm{}0.9$ \\
\bottomrule
\end{tabular}
\end{center}
\caption{Encoder model results by language on UDPOS Part-of-Speech tagging (POS) test sets, f1 score.}
\label{table:udpos_full}
\end{table*}

\begin{table*}[h]
\small
\begin{center}
\begin{tabular}{lc|cccc|c}
\toprule
Model & en & fr & de & es & hi & avg-0s\\ 
\midrule
\multicolumn{7}{c}{Seq2Seq Models From Scratch (de-noising only)} \\
\midrule
 bart-12e12d                & \textbf{83.4} $\pm0.2$ & \underline{54.3} $\pm1.2$ & 48.5 $\pm1.7$ & 51.6 $\pm1.6$ & 28.4 $\pm0.5$ & 45.7 $\pm1.1$ \\
 bart-12e12d-mask           & 83.2 $\pm0.5$ & 53.9 $\pm0.6$ & \underline{51.0} $\pm0.4$ & 53.2 $\pm0.9$ & 29.3 $\pm0.2$ & \underline{46.9} $\pm0.5$ \\
  \midrule
 \multicolumn{7}{c}{Recipe 2: Two-Stage Seq2Seq Models (warm-start with MLM encoder)} \\
\midrule
 2stage-bart-12e12d          & 82.0 $\pm1.1$ & 52.3 $\pm.06$ & 49.6 $\pm1.4$ & \underline{54.4} $\pm0.5$ & 28.8 $\pm0.3$ & 46.3 $\pm0.3$ \\
 2stage-bart-12e12d-attn-f  & 80.6 $\pm1.3$ & 52.6 $\pm0.7$ & 49.8 $\pm0.7$ & 53.7 $\pm0.8$ & \underline{29.7} $\pm0.4$ & 46.4 $\pm0.5$ \\
 2stage-bart-12e12d-unfrz         & \underline{83.3} $\pm0.2$ & \textbf{55.2} $\pm1.1$ & \textbf{51.3} $\pm1.6$ & \textbf{55.3} $\pm1.2$ & \textbf{31.1} $\pm0.2$ & \textbf{48.2} $\pm0.5$ \\
\bottomrule
\end{tabular}
\end{center}
\caption{Seq2Seq model results by language on mTOP semantic parsing test sets, SCIEM.
}
\label{table:mtop_full}
\end{table*}

\section{Fine-tuning Hyperparameters}
\label{sec:ft_hyperparams}

Table \ref{tab:hyperparams_enc} shows the hyperparameters for fine-tuning the pre-trained models.
For encoders, we first performed a single run with learning rates among 1e-6, 3e-6, 1e-5, 3e-5, 1e-4 for each task
and model, and found that the best learning rate was nearly always 1e-5 or 3e-5,
with only small differences between those two options by model.
For consistency, we then fixed the learning rate for each task
and ran each model on each task with three random seeds.
We use Adam \citep{Kingma2015AdamAM} optimization.
We freeze the embedding layer which we find generally slightly
improves the cross-lingual zero-shot results.

For XNLI, we follow the standard practice established in BERT \citep{devlin-etal-2019-bert}
to attach the classification head
to the first token (``\texttt{<s>}'' for all of our models).
We also explored max pooling across all tokens and did not observe
a significant difference in performance.

For mATIS++, following \citet{chen2019bert},
we use two separate classification heads, one for Intent Classification (IC) attached
to the encoder output of the first subword token of the sequence,
and the second for Slot Labeling (SL) attached to the \textit{first subword of each whole word}
in the sequence.

Similarly, for WikiANN NER and UDPOS, we again use a single classification head attached to the first subword of
each whole word in the sequence.
When computing f1 score for sequence labeling tasks (mATIS++ SL and WikiANN NER), we ignore the ``O'' (``Outside'')
tag, using the seqeval \citep{seqeval} implementation
which takes into account the BIO tags present in WikiANN.

\begin{table}[h]
\footnotesize
\setlength{\tabcolsep}{3pt}
\centering
\begin{tabular}{l|cccc|cc}
\toprule
 \multirow{2}{*}{Parameter}                       & \multicolumn{4}{c|}{Encoder tasks} & \multicolumn{2}{c}{Seq2Seq tasks}   \\
                        & XNLI          & mATIS++                           & WikiANN     & UDPOS  & mTOP        & XSUM   \\
\midrule
 Peak Learning Rate (LR)             & 1e-5          & 3e-5                              & 3e-5 &  3e-5          & 5e-6 & 5e-6 \\
 LR warmup type      & \makecell{linear \\ from 0} & \makecell{linear \\ from 0} & \makecell{linear \\ from 0} & \makecell{linear \\ from 0} & \makecell{exponential \\ from 1e-7} & \makecell{exponential \\ from 1e-7} \\
 LR warmup num steps & 1000          & 500                               & 300 & 1000            & 1000 & 1000 \\
 LR decay type       & linear to 0   & linear to 0                       & linear to 0 & linear to 0   & linear to 1e-7 & linear to 1e-7 \\
 Batch size                     & 128           & 128                               & 128 & 128           & 32 & 32 \\
 Epochs                         & 5             & 200                               & 20 & 56            & 200 & 200 \\
 Validation Metric & Accuracy             & \makecell{Slot Labeling f1}                               & \makecell{Slot Labeling f1} & \makecell{Slot Labeling f1}            &  Exact Match & Perplexity \\
 Max number of updates        & 30k           & 7k                                & 3k & 9k            &  $\sim$50k & $\sim$50k \\
 Classification head(s)         & [512] gelu    & \makecell{[256,256] gelu \\ each for \\ IC and SL} & [512] gelu & [512] gelu    & -- & -- \\
\bottomrule
\end{tabular}
\caption{Hyperparameters for fine-tuning. All models use AdamW with betas (0.9, 0.99), weight decay 0.1, and dropout 0.1.
For each run, we select the checkpoint with the best value of the target metric on the validation set.}
\label{tab:hyperparams_enc}
\end{table}

\section{Details on Compute Cost}
\label{sec:compute_cost_details}

We provide details on the compute cost reported in Table \ref{table:models}.
The unit ``TU'' (Training Updates) is defined as the compute cost for 100k updates
(forward and backward pass) of 12 model layers with hidden dimension 1024
and batch size 1M tokens.
The encoder-only MLM model trains for 500k updates, for Compute Cost 5.0 TU.
The Seq2Seq Models From Scratch have more layers, and therefore a larger
Compute Cost for 500k updates.
For example, ``bart-12e12d'' has 12 layers each for encoder and decoder,
resulting in a compute cost of 10.0 for 500k updates.
As a baseline, training both the MLM encoder
and the seq2seq models from scratch would incur a compute cost of
$5.0 + 10.0 = 15.0$ TU.

Recipe 1 (Encoder of Seq2Seq + MLM), first pays compute cost 10.0 TU for
the seq2seq training, then 1.0 TU for 100k MLM updates on the extracted encoder, for a total of 11.0 TU.

For Recipe 2 (Two-Stage Seq2Seq Models warm-started with MLM encoder),
we first pay a compute cost of 5.0 TU from MLM
pre-training of the encoder, then add compute cost for the second stage seq2seq pre-training.
When the encoder is frozen,
we only need to compute the forward pass for the encoder, not the backward pass.
We estimate that when the encoder is frozen, its forward pass uses $1/2$ the compute as
a forward and backward pass would use. (In reality, the ratio is likely less, as we also save memory by
not needing to store the optimizer states.)
Therefore, when the encoder is frozen for ``2stage-bart-12e12d'' and ``2stage-bart-12e12d-attn-f'',
the 500k decoder updates incur a compute cost of 2.5 on the encoder side
and 5.0 on the decoder side.
Adding this to the 5.0 for MLM initialization gives a total compute cost of $5.0 + 7.5 + 12.5$ TU.

For ``2stage-bart-12e12d-unfrz'', the 200k updates with frozen encoder incur a compute cost
of 1.0 TU on the encoder side and 2.0 TU on the decoder size for a total of 3.0 TU.
During the final 150k updates, the encoder is unfrozen, so the
compute cost is 3.0.
Adding the 5.0 compute cost for MLM Encoder initialization,
the total compute cost for this model is $5.0+3.0+3.0=11.0$ TU.

\section{Pre-Training Details}
\label{sec:pre_train_details}

\begin{figure*}[h]
\centering
\begin{subfigure}{0.32\textwidth}
  \centering
  \includegraphics[width=0.8\textwidth]{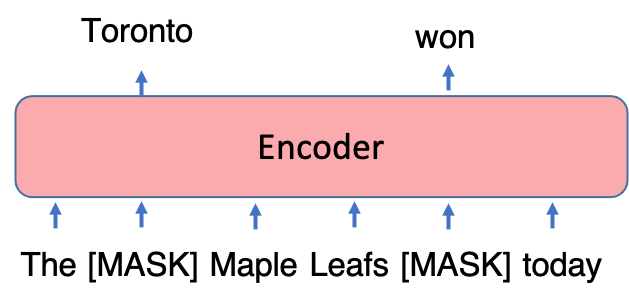}
\caption{
Masked Language Modeling (MLM).
}
  \label{fig:mlm}
\end{subfigure}%
\hfill
\begin{subfigure}{.32\textwidth}
  \centering
\includegraphics[width=0.8\textwidth]{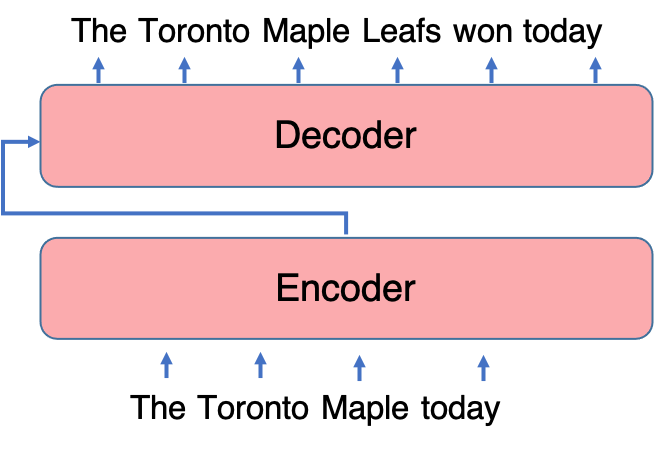}
\caption{
Seq2Seq de-noising.
}
  \label{fig:de_noising}
\end{subfigure}
\hfill
\begin{subfigure}{.32\textwidth}
  \centering
\includegraphics[width=0.8\textwidth]{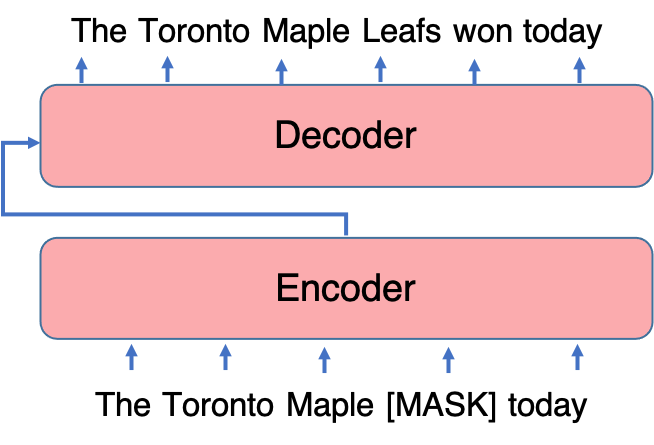}
\caption{
Seq2Seq de-noising with masking.
}
  \label{fig:de_noising_with_mask}
\end{subfigure}
\caption{
The training objectives we explore, with the example sentence
``The Toronto Maple Leafs won today''.
}
\label{fig:objectives}
\end{figure*}

We show an example sentence for each of our pre-training objectives
in Figure \ref{fig:objectives}.

Models were pre-trained (8 or 16 machines) and fine-tuned (1 machine) on AWS p3dn.24xlarge instances.
For MLM pre-training, we use a peak learning rate of 1.5e-4 (1e-4 for the second stage of Recipe 1)
warmed up over 5k update steps (1k for the second stage of Recipe 1) and decayed linearly
down to 5e-6 over the total number of updates (500k or 100k, respectively).
For seq2seq pre-training, we use the same learning rate as MLM pre-training:
peak of 1.5e-4, warmed up over 5k updates, and linearly decayed down to 5e-6 for the duration of pre-training.
For all pre-training runs, we use dropout of 0.1.

Our code is derived from HuggingFace \citep{wolf-etal-2020-transformers}.
We use DeepSpeed \citep{Rasley2020DeepSpeed} ZeRO Stage 1 to accelerate training.

\section{Dataset Sources}
\label{sec:dataset_source}

We show in Table \ref{tab:dataset_sources} the source locations of
the datasets we use for fine-tuning evaluation.

\begin{table}[h]
\footnotesize
\centering
\begin{tabular}{ll}
\toprule
 Dataset   & Source                                      \\
\midrule
 XNLI      & https://huggingface.co/datasets/xnli        \\
 mATIS++   & https://github.com/amazon-science/multiatis \\
 WikiANN   & https://huggingface.co/datasets/wikiann     \\
 UDPOS     & https://huggingface.co/datasets/xtreme      \\
 mTOP      & https://fb.me/mtop\_dataset                  \\
 XSUM      & https://huggingface.co/datasets/xsum        \\
\bottomrule
\end{tabular}
\caption{Source Locations for the fine-tuning datasets we evaluate on.}
\label{tab:dataset_sources}
\end{table}

\end{document}